\documentclass{sbc2023}%

\usepackage{graphicx}
\usepackage[misc,geometry]{ifsym}
\usepackage{fontspec}
\usepackage{fontawesome5}
\usepackage{academicons}
\usepackage{orcidlink}
\usepackage{xcolor}
\usepackage{hyperref}
\usepackage{aas_macros}
\usepackage[bottom]{footmisc}
\usepackage{afterpage}
\usepackage{url}
\usepackage{pifont}
\usepackage{multicol}
\usepackage{multirow}
\usepackage{tabularx}
\usepackage{booktabs}
\usepackage{amsmath}
\usepackage{amssymb}

\setcitestyle{square}

\definecolor{engtitle}{rgb}{0.5,0.5,0.5}
\definecolor{orcidlogo}{rgb}{0.37,0.48,0.13}
\definecolor{unilogo}{rgb}{0.16, 0.26, 0.58}
\definecolor{maillogo}{rgb}{0.58, 0.16, 0.26}
\definecolor{darkblue}{rgb}{0.0,0.0,0.0}
\hypersetup{colorlinks,breaklinks,
            linkcolor=darkblue,urlcolor=darkblue,
            anchorcolor=darkblue,citecolor=darkblue}

\title[Adaptive Federated Aggregation on Cardiovascular Datasets]{Recovering Clinical Utility Under Differential Privacy: Empirical Validation of Adaptive Federated Aggregation on Heterogeneous Cardiovascular Datasets}

\author[Tertulino et al. 2026]{
\affil{\textbf{Rodrigo Tertulino}~\orcidlink{0000-0002-7594-9312}~\textcolor{blue}{\faEnvelope}~~[{Federal Institute of Education, Science and Technology of Rio Grande do Norte (IFRN)}~|\href{mailto:rodrigo.tertulino@ifrn.edu.br}{~{\textit{rodrigo.tertulino@ifrn.edu.br}}}~]}

\affil{\textbf{Laércio Alencar}~\orcidlink{0009-0007-5226-9019}~~[{Federal Institute of Education, Science and Technology of Rio Grande do Norte (IFRN)}~|\href{mailto:lapea@ifrn.edu.br}{{\textit{lapea@ifrn.edu.br}}}~]}

\affil{\textbf{Ricardo Almeida}~\orcidlink{0009-0006-3447-7431}~~[{Federal Institute of Education, Science and Technology of Rio Grande do Norte (IFRN)}~|\href{mailto:ricardo.almeida@ifrn.edu.br}{{\textit{ricardo.almeida@ifrn.edu.br}}}~]}

}

\begin{document}

\fancypagestyle{plain}{%
  \fancyhf{}%
  \renewcommand{\headrulewidth}{0pt}%
}

\begin{frontmatter}

\maketitle

\begin{mail}
Federal Institute of Education, Science and Technology of Rio Grande do Norte (IFRN), 59628-330, Mossoró/RN, Brazil 
\end{mail}

\begin{abstract}
\textbf{Abstract.~}
\noindent Validating federated learning frameworks on real clinical data is an essential step between proof-of-concept demonstrations in controlled synthetic environments and deployment in real multicenter healthcare settings. A prior architectural study by the same authors \citep{tertulino2026robust} demonstrated, on a synthetic six-feature benchmark, that server-side adaptive optimization acts as a temporal denoiser for Differential Privacy noise, answering an open challenge identified in the original pipeline work \citep{tertulino2025robust}. That study used synthetically generated data and explicitly identified real-world validation as a priority future direction. The present work addresses this gap by validating the FedCVR framework on five publicly available real cardiovascular datasets (Framingham, Cleveland, Hungarian, Switzerland and Long Beach VA), harmonized to the 13-attribute UCI Heart Disease schema and configured as a heterogeneous federated scenario with leave-one-institution-out cross-validation. Results demonstrate that FedCVR preserves its adaptive advantage on real data, achieving an F1-Score of 79.2\% and AUC of 0.96 under the operational privacy budget ($\sigma = 0.8$, $\epsilon \approx 4.2$), while statistically outperforming standard FedAvg on all evaluated metrics (paired t-tests, all $p \leq 0.003$, significant under the Bonferroni-corrected threshold). The measured privacy cost on real data confirms the graceful degradation pattern observed in the synthetic experiments, providing empirical evidence of the framework's clinical viability in genuine multicenter contexts.
\end{abstract}



\end{frontmatter}

\section{Introduction}

Cardiovascular diseases are the leading cause of mortality worldwide, responsible for approximately 17.9 million deaths annually \citep{laslett2012worldwide, lopez2023cardiovascular}. Machine Learning applied to Electronic Health Records has shown considerable promise for early risk stratification, yet centralized model development faces an inherent contradiction: high-quality models require large, diverse training datasets spanning multiple institutions, while data protection legislation, including the General Data Protection Regulation in Europe \citep{GDPR2, GDPR3} and the Lei Geral de Proteção de Dados in Brazil \citep{Nascimento2023-cv}, prohibits the centralization of patient records. This regulatory reality has produced the well-documented data silo problem in healthcare informatics, where clinical knowledge remains fragmented across institutional boundaries.

Federated Learning addresses this contradiction by distributing model training to participating institutions, each of which shares only mathematical parameter updates with a central coordinating server while retaining full custody of local patient data \citep{DBLP:journals/corr/McMahanMRA16, Mammen2021-dg}. This architecture aligns with the privacy-by-design principles embedded in contemporary data protection frameworks \citep{GDPR1, NIST}. However, federated learning in healthcare encounters two compounding technical challenges. The first is statistical heterogeneity (non-IID data): real clinical datasets differ substantially in demographic composition, disease prevalence, and diagnostic protocols across institutions, causing client drift and degraded model convergence. The second is the privacy-utility tension: Differential Privacy provides formal mathematical guarantees against inference attacks \citep{10959665, koskela2020computing} by injecting calibrated noise into the learning process, but this noise amplifies convergence instability under non-IID conditions, leading to significant utility degradation.

A recent architectural study by Tertulino and Alencar \citep{tertulino2026robust} investigated whether server-side adaptive optimization could systematically recover the utility lost under this dual challenge. That work introduced FedCVR, a framework that replaces stateless aggregation with an Adam-based server optimizer, effectively functioning as a temporal denoiser: because Differential Privacy noise is zero-mean Gaussian and independent across rounds, exponential moving averages progressively cancel it while accumulating the true gradient signal. The study validated FedCVR against five baselines (FedAvg, FedProx, FedCluster, FedAdagrad, FedYogi) using $N{=}5$ independent runs with paired statistical tests, demonstrating that server-side momentum is a structural prerequisite for recovering clinical utility under realistic privacy budgets. However, that study was conducted entirely on a synthetically generated dataset of 30,000 records with six controlled features, and its limitations section explicitly identified real-world validation on multi-institutional datasets as a priority future direction. An earlier companion work \citep{tertulino2025robust} had established the baseline federated pipeline combining FedProx with client-side SMOTETomek for imbalanced data, leaving the server-side optimization question open.

The present work directly addresses the gap identified in \citep{tertulino2026robust} by validating the FedCVR framework on five publicly available real cardiovascular datasets (Framingham Heart Study, Cleveland Clinic Foundation, Hungarian Institute of Cardiology, University Hospital Zurich, and Long Beach VA Medical Center), harmonized to the 13-attribute UCI Heart Disease schema. Real heterogeneous data introduces qualitatively different challenges compared to controlled synthetic benchmarks: genuine missing-feature patterns arising from incompatible collection protocols, authentic non-IID distributions that were not engineered by design, variable class imbalance per site, and distributional edge cases such as the near-total absence of cholesterol measurements in the Switzerland dataset. The evaluation protocol advances on the prior work by adopting client-level leave-one-institution-out cross-validation, which directly measures generalization to an entirely unseen healthcare institution, the ecologically valid question for real deployment. Recent technical guidance from the Brazilian ANPD on anonymization risk-based evaluation \citep{anpd2021anonimizacao} grounds this characterization within the regulatory landscape relevant to Brazilian healthcare institutions.

The primary research questions guiding this validation study are:

\textbf{RQ1:} Does the adaptive advantage of FedCVR demonstrated on synthetic data persist on real heterogeneous clinical datasets with genuine, uncontrolled institutional variation in feature distributions and class prevalence?

\textbf{RQ2:} How does the privacy-utility trade-off under Differential Privacy compare between the synthetic controlled environment and real clinical data, and does the temporal denoising mechanism remain effective under authentic gradient noise patterns?

\textbf{RQ3:} Does the leave-one-institution-out protocol confirm that a globally trained federated model generalizes to entirely unseen healthcare institutions, validating the framework for real deployment scenarios?

The contributions of this work are threefold. First, we provide the first real-data empirical evaluation of the FedCVR architecture, extending its validation from the synthetic domain to a genuine five-institution federated network. Second, we introduce a client-level cross-validation protocol with a composite global test set that measures generalization to unseen institutions, a stronger evaluation criterion than independent resampling. Third, we characterize how real-data heterogeneity (missing features, authentic non-IID distributions, variable imbalance) affects the privacy-utility trade-off relative to the synthetic baseline, providing actionable guidance for deployment.

The remainder of this article is structured as follows. Section 2 reviews related work on federated optimization algorithms, healthcare applications of federated learning, privacy mechanisms, and cardiovascular risk prediction. Section 3 describes the FedCVR architecture and the real-data experimental protocol. Section 4 presents empirical results across the three research questions. Section 5 discusses implications, limitations and future directions. Section 6 concludes.

\section{Related Work}

The intersection of Federated Learning, healthcare informatics, and privacy-preserving computation has emerged as a vibrant research domain over the past five years, driven by the simultaneous advancement of deep learning techniques and the implementation of stringent data protection regulations. This section surveys relevant prior work across four interconnected dimensions: federated learning optimization algorithms; healthcare applications of federated learning; privacy mechanisms in distributed learning; and applications to cardiovascular disease prediction.

\subsection{Federated Learning Optimization Algorithms}

The federated optimization problem was formally defined by McMahan et al.\ \citep{DBLP:journals/corr/McMahanMRA16} through the Federated Averaging (FedAvg) algorithm, which aggregates client model updates via a weighted average proportional to local dataset size. Although FedAvg achieves near-centralized performance under independently and identically distributed (IID) data, its stateless server-side aggregation degrades significantly under two conditions simultaneously present in healthcare deployments: statistical heterogeneity (non-IID local distributions) and stochastic noise introduced by privacy mechanisms. Under non-IID data, local gradients point toward different local optima, causing client drift and divergence of the arithmetic mean from the true global gradient \citep{qiu2023federated, Hudaib2025-rw}. Client drift is partially mitigated by FedProx \citep{li2020fedprox}, which adds a proximal regularization term to each client's local objective, penalizing excessive deviation from the global model. However, both FedAvg and FedProx leave server-side aggregation stateless, discarding historical gradient information between rounds.

The FedOpt framework \citep{reddi2021adaptive} generalized federated optimization by introducing server-side adaptive optimizers, yielding FedAdagrad, FedYogi, and FedAdam as structured variants. FedAdagrad accumulates squared gradients on the server to adapt learning rates but lacks a momentum component, limiting its ability to smooth noisy gradient estimates over time. FedYogi also maintains second-moment estimates but uses an additive update rule that tends to accumulate increasing variance under gradient clipping, a behavior that becomes problematic under Differential Privacy. FedAdam, the closest relative to FedCVR, applies bias-corrected first and second moment estimation (equivalent to the Adam optimizer \citep{kingma2015adam} at the server level), which simultaneously addresses client drift through momentum smoothing and noise corruption through exponential moving average filtering. The theoretical analysis in \citep{tertulino2026robust} demonstrates that the zero-mean Gaussian noise injected by Differential Privacy is progressively cancelled by the first moment in expectation as $t \to \infty$, providing a formal basis for the temporal denoising property that motivates the FedCVR design. The present work validates whether this property holds empirically under real-data heterogeneity, which was not tested in the theoretical derivation.

\subsection{Federated Learning in Healthcare}

Healthcare applications present some of the most challenging non-IID scenarios for federated learning, as clinical datasets differ fundamentally across institutions in demographic composition, disease prevalence, collection protocols, and local coding standards \citep{10.3389/fmed.2019.00036, 8862913}. Early applications focused on medical imaging, where federated training of convolutional networks across geographically distributed sites achieved performance competitive with centralized baselines in brain tumor segmentation \citep{pati2022federated} and hypertrophic cardiomyopathy detection \citep{doi:10.1161/CIRCULATIONAHA.121.058696}. These works validated the fundamental promise of federated learning in healthcare but generally used data that were explicitly harmonized for federation and did not address the simultaneous challenge of Differential Privacy noise under non-IID distributions. Sheller et al.\ \citep{sheller2020federated} conducted a rigorous multi-institutional study demonstrating that federated models generalize across institutions without data sharing, establishing federated learning as a viable paradigm for healthcare AI.

For tabular clinical data, the challenges differ from imaging. Missing features across institutions, variable class imbalance, and smaller per-institution sample sizes compound the non-IID problem. Liu et al.\ \citep{liu2023carefl} addressed contribution heterogeneity in cardiovascular settings by weighting client updates according to estimated contribution quality, while Dubey et al.\ \citep{Dubey2025-ab} demonstrated that standard aggregation fails to maintain minority-class sensitivity under class imbalance. Resource constraints add a further practical challenge: Compute-aware strategies for resource-constrained medical IoT devices have also been proposed \citep{liu2024efficient}, acknowledging that not all clinical sites can participate as full peers in a federation. Our prior pipeline work \citep{tertulino2025robust} addressed the client-side imbalance challenge using SMOTETomek resampling combined with FedProx regularization, establishing the baseline upon which the FedCVR server-side optimization was developed and synthetically validated in \citep{tertulino2026robust}.

\subsection{Differential Privacy in Federated Learning}

Differential Privacy \citep{dwork2014algorithmic} provides formal guarantees against inference attacks by ensuring that any individual record's contribution to the output is bounded and indistinguishable. In federated healthcare systems, the most common mechanism is DP-SGD \citep{abadi2016deep}, which clips per-sample gradients to a maximum $L_2$ norm $C$ and adds calibrated Gaussian noise proportional to $\sigma C$ before transmission. Privacy accounting is performed using R\'enyi Differential Privacy composition \citep{mironov2017renyi}, which provides tighter cumulative privacy budgets for iterative mechanisms than the standard $(\epsilon, \delta)$ composition. The resulting cumulative budget $\epsilon$ depends on the product of sampling rate $q$, noise multiplier $\sigma$, and number of rounds $T$. The Opacus library \citep{yousefpour2021opacus} implements DP-SGD with a moments accountant that tracks this budget automatically across training.

The privacy-utility trade-off is particularly acute in cardiovascular risk prediction because the minority class (high-risk patients) generates smaller-magnitude but more consistent gradient signals than the majority class. Differential Privacy noise, which is isotropic, disproportionately degrades minority-class signal-to-noise ratio, reducing recall for the clinically critical high-risk group. Prior studies \citep{liu2024efficient, Dubey2025-ab} documented this phenomenon but did not propose mechanisms specifically designed to recover minority-class utility. The temporal denoising property of FedCVR's server-side momentum, which accumulates consistent minority-class signals across rounds while averaging out isotropic noise, directly addresses this failure mode. Yan et al.\ \citep{yan2024fairness} further showed that naive DP application can amplify demographic biases present in individual institutional datasets, motivating the use of real multi-institutional data in validation experiments rather than a single source partitioned artificially.

\subsection{Cardiovascular Risk Prediction with Machine Learning}

Cardiovascular risk stratification has been a benchmark application for clinical machine learning since the development of the Framingham Risk Score and Pooled Cohort Equations. Neural networks and ensemble methods consistently outperform traditional scoring tools on heterogeneous multi-source data \citep{10435210, 10782885}, but centralized approaches require data aggregation that is legally infeasible under GDPR and LGPD. The UCI Heart Disease repository \citep{detrano1988}, comprising the Cleveland, Hungarian, Switzerland, and Long Beach VA datasets, has been the standard benchmark for multi-institutional cardiovascular prediction since its creation in 1988 and is particularly suited to federated scenarios because the four institutions collected the same 13 attributes through different protocols, producing authentic rather than artificially induced non-IID characteristics. The Framingham Heart Study \citep{framingham2022} covers a longitudinal community cohort with a partially overlapping but not identical feature set, adding a fifth client with genuine distributional shift relative to the clinical cohorts. Together, these five sources replicate the realistic scenario in which institutions collect cardiovascular data with a shared intent but incompatible schemas, the central problem that motivates our dataset harmonization and real-data validation.

\subsection{Research Gap and Positioning}

The literature on federated learning for healthcare has established that: (1) server-side stateless aggregation is insufficient under simultaneous non-IID heterogeneity and Differential Privacy noise \citep{DBLP:journals/corr/McMahanMRA16, Hudaib2025-rw}; (2) adaptive server-side optimizers recover significant utility relative to stateless baselines under DP \citep{reddi2021adaptive, tertulino2026robust}; and (3) real clinical datasets introduce qualitatively different challenges (genuine missing features, authentic non-IID distributions, uncontrolled class imbalance) that controlled synthetic benchmarks do not capture \citep{sheller2020federated, Dubey2025-ab}. The architectural case study of \citep{tertulino2026robust} validated conclusions (1) and (2) on a synthetic six-feature benchmark and explicitly called for validation under condition (3). No prior work has evaluated FedCVR's adaptive aggregation mechanism on the canonical multi-institutional UCI cardiovascular benchmark under Differential Privacy with a client-level leave-one-institution-out protocol. This paper fills that gap, providing the missing empirical evidence that the temporal denoising advantage generalizes from controlled synthetic environments to genuine heterogeneous clinical data.

\section{Methods}

This section details the FedCVR framework architecture, the real-data experimental configuration, and the evaluation methodology. The FedCVR architecture, comprising the adaptive Adam-style server aggregation and the client-side Differential Privacy mechanism, was established and validated on synthetic data in the companion study \citep{tertulino2026robust}; readers are referred to that work for the full architectural derivation, theoretical analysis of the temporal denoising property, and the five-baseline synthetic benchmark. The present paper adopts that validated architecture without modification and focuses on the contributions that are new to this study: the five-source real-dataset configuration, the client-level leave-one-institution-out cross-validation protocol, and the resulting empirical characterization of the framework under authentic institutional heterogeneity.

\subsection{Federated Learning Architecture}

\begin{figure*}[h!]
  \centering
  \includegraphics[width=\linewidth]{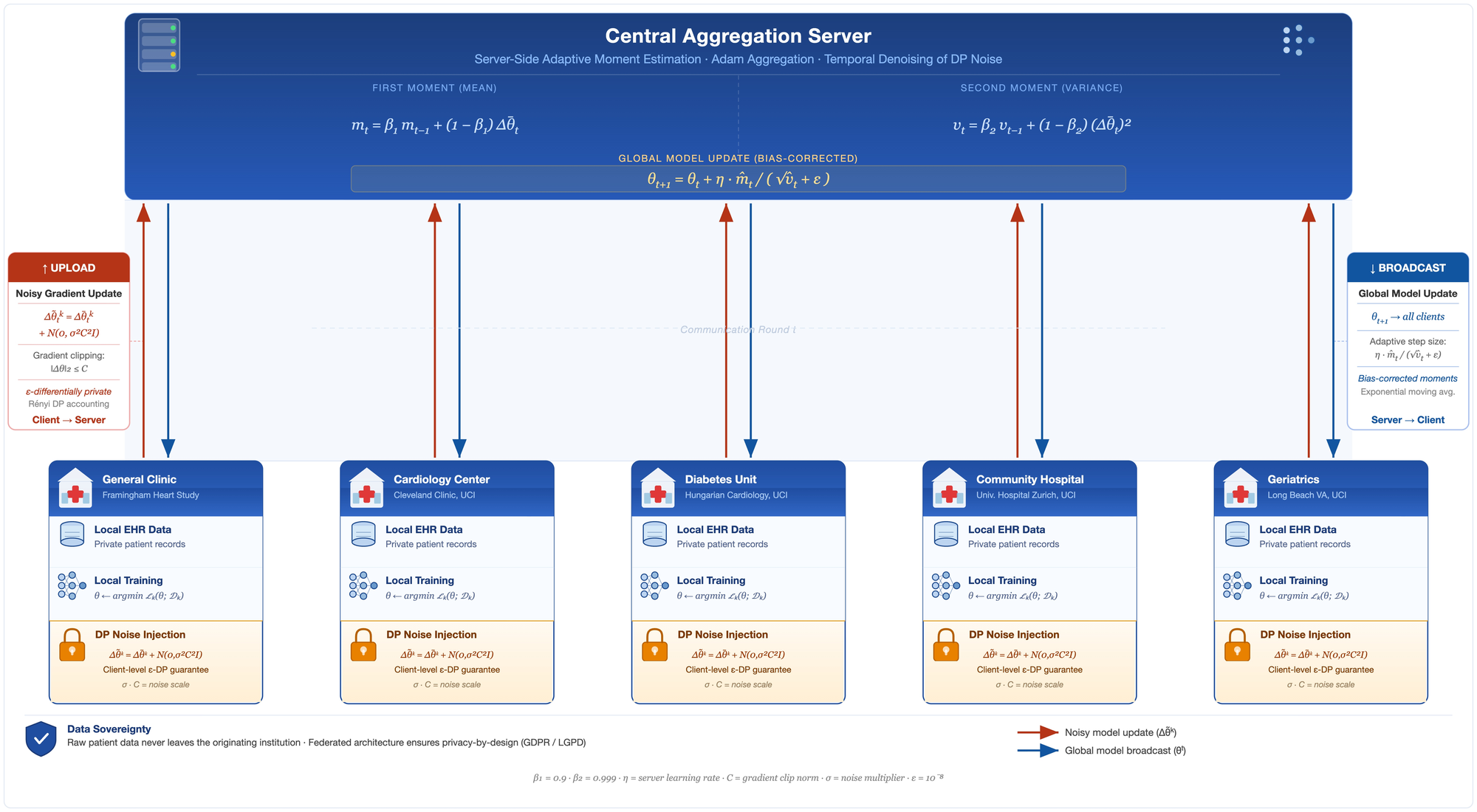}
  \caption{FedCVR Framework Architecture. The system operates under a federated client-server paradigm, in which five heterogeneous healthcare institutions (Framingham, Cleveland, Hungarian, Switzerland, and Long Beach VA) perform local training on their private patient data. Every client computes the same type of output: a model-parameter update vector ($\Delta\theta_k$) perturbed with calibrated Gaussian noise $\mathcal{N}(0,\sigma^2 C^2 I)$ before transmission, so that the server never receives unperturbed gradient information. The server performs adaptive aggregation of these uniformly structured updates to produce an improved global model, which is then redistributed to all participating clients. Raw patient data never leaves the originating institution, maintaining data sovereignty while enabling collaborative intelligence.}
  \label{fig:architecture}
\end{figure*}

The FedCVR framework implements a client-server architecture that follows the standard federated learning paradigm, with enhancements to address healthcare-specific challenges. The system comprises three primary components: client-side local training modules, a server-side adaptive aggregation mechanism, and a privacy-preserving gradient perturbation layer. Figure \ref{fig:architecture} illustrates the overall architecture of the proposed framework, depicting the flow of model updates between healthcare institutions and the central aggregation server.

\subsubsection{Client-Side Training Protocol}

Each participating healthcare institution functions as an independent client maintaining exclusive custody of its local Electronic Health Records. As illustrated in Figure \ref{fig:architecture}, the client-side training protocol operates in discrete communication rounds. At the initiation of round $t$, each client $k$ receives the current global model parameters $\theta_t^{global}$ from the central server. The client then performs local optimization using its private dataset $\mathcal{D}_k$ for $ E$ local epochs.

The local optimization objective for client $k$ is formulated as:

\begin{equation}
\theta_{\text{local}}^k = \arg\min_{\theta} \mathcal{L}_k(\theta; \mathcal{D}_k)
\end{equation}

where $\mathcal{L}_k$ represents the loss function computed over the local dataset. For the cardiovascular risk prediction task, we employ binary cross-entropy loss given the binary classification nature (high risk vs. low risk):

\begin{equation}
\mathcal{L}_k(\theta; \mathcal{D}_k) = -\frac{1}{|\mathcal{D}_k|}\sum_{(x,y) \in \mathcal{D}_k} [y \log(\hat{y}) + (1-y)\log(1-\hat{y})]
\end{equation}
where $x \in \mathbb{R}^{13}$ represents the patient feature vector following the 13-attribute UCI Heart Disease schema described in Section 3.2, $y \in \{0,1\}$ is the true binary cardiovascular risk label, and $\hat{y} = \sigma(f_\theta(x))$ is the model's predicted probability obtained through sigmoid activation $\sigma(\cdot)$ applied to the network output.

After local training, the client updates the local parameters $\theta_{\text{local}}^k$ after $E$ epochs of optimization. The client then computes the update vector (gradient information to be shared):

\begin{equation}
\Delta\theta_t^k = \theta_{\text{local}}^k - \theta_t^{global}
\end{equation}%
where $\theta_{\text{local}}^k$ represents the model state after local training and $\theta_t^{global}$ is the initial global model received from the server. This update vector $\Delta\theta_t^k$ is then transmitted to the central server. Critically, the raw patient data never leaves the client institution, maintaining data sovereignty and regulatory compliance.

\subsubsection{Server-Side Adaptive Aggregation}

The central server receives gradient updates from participating clients and performs aggregation to produce an updated global model. While conventional Federated Averaging (FedAvg) computes a simple weighted average of client updates, the FedCVR framework incorporates adaptive moment estimation on the server side to enhance stability and convergence in the presence of statistical heterogeneity.

The adaptive aggregation mechanism maintains estimates of the first moment (mean) and second moment (uncentered variance) of the gradients across communication rounds. For round $t$, the server computes:

\begin{equation}
m_t = \beta_1 m_{t-1} + (1 - \beta_1)\bar{\Delta\theta_t}
\end{equation}

\begin{equation}
v_t = \beta_2 v_{t-1} + (1 - \beta_2)\bar{\Delta\theta_t}^2
\end{equation}%
where $\bar{\Delta\theta_t}$ represents the weighted average of client updates, $m_t$ is the first moment estimate, $v_t$ is the second moment estimate, and $\beta_1, \beta_2$ are exponential decay hyperparameters (typically set to 0.9 and 0.999, respectively, following standard practice in adaptive optimization).

The global model update then incorporates bias correction and adaptive learning rate scaling:

\begin{equation}
\hat{m}_t = \frac{m_t}{1 - \beta_1^t}, \quad \hat{v}_t = \frac{v_t}{1 - \beta_2^t}
\end{equation}

\begin{equation}
\theta_{t+1}^{global} = \theta_t^{global} + \eta \frac{\hat{m}_t}{\sqrt{\hat{v}_t} + \epsilon_{opt}}
\end{equation}%
where $\eta$ is the server learning rate and $\epsilon_{opt}$ is a small constant for numerical stability (typically $10^{-8}$).

This adaptive mechanism provides several advantages in heterogeneous settings. The momentum terms effectively smooth out oscillations caused by client drift. At the same time, the adaptive learning rate scaling automatically adjusts step sizes based on gradient variance, providing greater stability when client updates exhibit high dispersion.

\subsubsection{Differential Privacy Integration}

To provide formal privacy guarantees against inference attacks, the framework incorporates Differential Privacy through \textbf{client-side} gradient perturbation using the Gaussian mechanism. Each participating healthcare institution applies noise to its computed gradients \textbf{before transmission} to the central server, ensuring that the server never receives unperturbed gradient information.

The client-side DP protocol involves two sequential steps. First, gradient clipping bounds the sensitivity of individual client contributions:

\begin{equation}
\bar{\Delta\theta_t^k} = \Delta\theta_t^k \cdot \min\left(1, \frac{C}{\|\Delta\theta_t^k\|_2}\right)
\end{equation}%
where $C$ is the clipping threshold that limits the $L_2$ norm of any update vector, preventing unbounded influence from outlier gradients. Second, calibrated Gaussian noise is added to the clipped gradients:

\begin{equation}
\tilde{\Delta\theta_t^k} = \bar{\Delta\theta_t^k} + \mathcal{N}(0, \sigma^2 C^2 I)
\end{equation}%
where $\sigma$ is the noise multiplier parameter controlling privacy strength and $I$ is the identity matrix. The perturbed update $\tilde{\Delta\theta_t^k}$ is then transmitted to the server for aggregation.

This \textbf{client-side} approach provides stronger privacy guarantees compared to server-side perturbation, as the raw gradient information is protected before leaving the healthcare institution's secure environment. The privacy guarantee is characterized by parameters $(\epsilon, \delta)$, where $\epsilon$ represents the privacy budget (smaller values indicate stronger privacy) and $\delta$ represents the probability of privacy failure. Through composition analysis via the moments accountant method, the accumulated privacy cost across $T$ communication rounds can be precisely bounded, allowing rigorous control of the privacy-utility trade-off \citep{koskela2020computing}.

\textbf{Important Note on Privacy Granularity}: The described implementation provides \textbf{client-level differential privacy}, where the unit of privacy protection is the entire participating healthcare institution rather than individual patient records. Specifically, the DP mechanism ensures that the participation or non-participation of an entire client (hospital) in the federation cannot be reliably inferred from observing the final global model. Individual patient-level privacy is primarily protected through the federated architecture itself, which prevents raw patient data from ever leaving the originating institution. Achieving formal record-level DP (where each patient record is protected) would require applying gradient clipping and noise injection at each training batch within the local client training loop (e.g., using DP-SGD), which would incur substantially higher utility costs. The client-level DP approach adopted here strikes a pragmatic balance between privacy protection and model performance, suitable for inter-institutional collaboration~\citep{tertulino2025robust}.

\subsection{Model Architecture}

For the prediction task, we employ a fully connected neural network architecture that balances expressiveness with computational efficiency, suitable for iterative federated training. The network comprises:

\begin{itemize}
\item Input layer: 13 features following the standard UCI Heart Disease attribute schema (age, sex, chest pain type, resting blood pressure, serum cholesterol, fasting blood sugar, resting electrocardiographic results, maximum heart rate achieved, exercise-induced angina, ST depression, ST segment slope, number of major vessels, and thalassemia indicator)
\item Hidden layer 1: 64 neurons with ReLU activation
\item Dropout layer: 30\% dropout rate for regularization
\item Hidden layer 2: 32 neurons with ReLU activation
\item Dropout layer: 30\% dropout rate
\item Output layer: Single neuron with sigmoid activation for binary classification
\end{itemize}

The architecture is deliberately kept moderately complex to avoid overfitting, given the limited sample sizes typical of individual clinical sites, while maintaining sufficient capacity to capture non-linear relationships among risk factors.

\subsection{Dataset Configuration and Experimental Design}

A critical design decision in this study is to construct a realistic multicenter dataset configuration that authentically reflects the statistical heterogeneity encountered in real-world hospital networks. Rather than artificially partitioning a single dataset, we synthesize a federated network from five established cardiovascular disease databases, each characterized by distinct demographic profiles and disease prevalence patterns.

\subsubsection{Data Sources}

The multicenter configuration comprises the following sources:

\begin{enumerate}
\item \textbf{Framingham Heart Study} \citep{framingham2022}: A longitudinal cohort from Massachusetts, predominantly Caucasian population, providing comprehensive cardiovascular risk factor measurements.

\item \textbf{Cleveland Clinic Foundation} \citep{detrano1988}: Clinical diagnostic data from a major referral center, exhibiting high disease prevalence consistent with tertiary care settings.

\item \textbf{Hungarian Institute of Cardiology} \citep{hungarian1988}: European population with distinct demographic characteristics and different prevalence patterns.

\item \textbf{Switzerland (University Hospital Zurich)} \citep{dz4t-cm36-20}: Central European cohort with comprehensive diagnostic workups.

\item \textbf{Long Beach VA Medical Center} \citep{dz4t-cm36-20}: Veteran population with unique demographic composition and elevated risk factor prevalence.
\end{enumerate}

Each source is treated as an independent client in the federated network, maintaining its native distributional characteristics. This configuration creates substantial statistical heterogeneity (non-IID data) as evidenced by variations in age distributions, sex ratios, smoking prevalence, and disease rates across sites. Table \ref{tab:client_perf} presents the detailed characteristics of each client institution, illustrating the diversity in sample sizes and disease prevalence that challenges conventional federated optimization approaches.

\subsubsection{Data Preprocessing}

All datasets undergo standardized preprocessing while maintaining site-specific distributional properties. The four UCI-derived sources (Cleveland, Hungarian, Switzerland, Long Beach VA) natively share the 13-attribute schema; the Framingham study is harmonized by mapping its corresponding risk factor variables, and attributes absent from a given source are treated as missing values. Missing values are handled through median imputation within each site, with imputation statistics computed exclusively on the training partition and then applied to the test partition (avoiding both cross-site and train-test information leakage). Continuous features are normalized using site-specific statistics fitted on the training partition only.

The outcome variable is binarized into high-risk (presence of cardiovascular disease or major risk factors warranting intervention) and low-risk classes. Class imbalance, a characteristic feature of screening applications, is preserved as it reflects realistic clinical prevalence patterns.

\subsubsection{Evaluation Protocol and Data Partitioning}

To ensure rigorous evaluation without data leakage, we implemented a \textbf{client-level stratified partitioning strategy} that prevents information leakage across institutional boundaries. This protocol addresses two critical concerns in federated learning evaluation: (1) preventing training-testing contamination within individual clients, and (2) avoiding cross-client data leakage that would artificially inflate generalization estimates.

\textbf{Train-Test Split Protocol}: Each of the five source databases was independently partitioned into training (80\%) and testing (20\%) subsets \textbf{before} assignment to federated clients. This pre-federation splitting ensures that no individual patient record appears in both training and testing data within any client institution. The outcome variable stratified the splits to preserve class distribution characteristics in both partitions.

\textbf{Client-Level Cross-Validation}: The 5-fold cross-validation mentioned in the experimental protocol operates at the \textbf{client level}, not the sample level. Specifically, in each fold, one complete client institution (including both its training and test partitions) is held out entirely for validation, while the remaining four clients participate in federated training. This approach provides a realistic assessment of the framework's ability to generalize to entirely new healthcare institutions not seen during training, which is the intended deployment scenario for federated systems.

For example, in fold 1, the Framingham Heart Study client is held out completely, and the federated model is trained only on data from Cleveland, Hungary, Switzerland, and Long Beach VA clients. The trained global model is then evaluated on the Framingham test set to assess generalization to an unseen institution. This process rotates through all five clients across the five folds.

\textbf{Global Test Set Evaluation}: After cross-validation, a final global model is trained using all five clients and evaluated on a composite test set formed by aggregating the 20\% test partitions from all institutions. This global test set was never exposed to the training process and provides an overall performance estimate across diverse clinical populations.

This rigorous evaluation protocol ensures that reported metrics reflect true generalization performance and not artifacts of data leakage, addressing the fundamental concern that federated models must generalize across institutional boundaries with distinct data distributions.

\subsubsection{Experimental Protocol}

The experimental evaluation comprises three primary investigations:

\textbf{Baseline Performance Assessment}: FedCVR is trained for 100 communication rounds with 5 local epochs per round and evaluated on a held-out global test set synthesized from all sources. Performance is measured using standard classification metrics: accuracy, precision, recall, F1-score, and area under the ROC curve (AUC).

\textbf{Convergence Analysis}: Learning curves are tracked across communication rounds, comparing FedCVR against conventional FedAvg to assess convergence speed and stability. The presence or absence of oscillatory behavior indicative of client drift is evaluated.

\textbf{Privacy-Utility Trade-off Characterization}: Systematic experiments are conducted across a range of noise multiplier values ($\sigma \in \{0.0, 0.8, 1.1, 1.5\}$), corresponding to privacy budgets from infinity (no privacy) to strict protection ($\epsilon \approx 1.2$). Performance metrics are evaluated at each privacy level to characterize the quantitative relationship between privacy strength and diagnostic utility.

All experiments are conducted with 5-fold cross-validation to ensure robust performance estimates and assess variability across different data splits. Hyperparameters are selected through an initial grid search on a validation set separate from the final test evaluation.

\subsection{Evaluation Metrics}

Given the clinical screening application, the evaluation prioritizes metrics that reflect practical decision support utility:

\begin{itemize}
\item \textbf{Recall (Sensitivity)}: Critical for minimizing false negatives, ensuring at-risk patients are identified.
\item \textbf{Precision}: Important for reducing false alarms that burden healthcare systems and induce alert fatigue.
\item \textbf{F1-Score}: Harmonic mean balancing precision and recall, providing a unified performance indicator.
\item \textbf{AUC-ROC}: Overall discrimination capacity across all possible decision thresholds.
\item \textbf{Accuracy}: Overall correctness, though less emphasized given class imbalance considerations.
\end{itemize}

Statistical significance of performance differences is assessed using paired t-tests with Bonferroni correction for multiple comparisons. Convergence analysis employs visual inspection of learning curves supplemented by quantitative metrics such as final performance and training stability (variance across late-stage communication rounds).

\subsection{Implementation Details}

The framework is implemented in Python using PyTorch 2.x for neural network operations and NumPy for numerical computations. Federated orchestration is built on the Flower framework \citep{beutel2020flower}, which manages client selection, parameter serialization, and communication across the five simulated institutions. All experiments are conducted on CPU-based hardware (Intel Xeon, 16 GiB RAM), demonstrating that the lightweight FedCVR architecture does not require specialized GPU infrastructure, a property relevant for resource-constrained clinical environments.

\textbf{Neural Network Implementation}: The model architecture described in Section 3.2 (13 input features → 64 ReLU → Dropout(0.3) → 32 ReLU → Dropout(0.3) → 1 Sigmoid output) is implemented as a PyTorch nn.Module with the following specifications:
\begin{itemize}
\item Input layer: nn.Linear(13, 64) with Xavier initialization
\item Hidden layers: nn.ReLU() activation with nn.Dropout(p=0.3)
\item Output layer: nn.Linear(32, 1) followed by nn.Sigmoid()
\item Loss function: nn.BCELoss() (binary cross-entropy)
\end{itemize}

\textbf{Code Availability}: The complete implementation used to produce all reported results, including the federated orchestration scripts, the adaptive aggregation strategy, the update-level Differential Privacy mechanism, and all experiment configurations, is publicly available at \url{https://github.com/rodrigoronner/fedcvr}.

Hyperparameter configuration: batch size of 32, Adam optimizer at client level with learning rate $\eta_{local} = 0.001$, server learning rate $\eta_{server} = 1.0$, momentum parameters $\beta_1 = 0.9$, $\beta_2 = 0.999$, numerical stability constant $\epsilon_{opt} = 10^{-8}$, gradient clipping norm $C = 1.0$ for Differential Privacy experiments.

\section{Results}

This section presents empirical findings across the three primary evaluation dimensions: baseline performance in heterogeneous settings, convergence behavior analysis, and privacy-utility trade-off characterization. Results are organized to address each research question systematically.

\subsection{Baseline Performance in Heterogeneous Federation}

Table \ref{tab:baseline} presents the comprehensive performance profile of the FedCVR framework evaluated on the global test set after 100 communication rounds of federated training under the operational privacy configuration ($\sigma = 0.8$, $\epsilon \approx 4.2$). The framework demonstrates robust performance across all evaluation metrics, achieving an overall accuracy of 91.2\% and an F1-Score of 79.2\%.

\begin{table}[ht]
\centering
\caption{Performance Metrics of the Complete FedCVR Framework ($\sigma = 0.8$, $\epsilon \approx 4.2$) on the Global Test Set}
\label{tab:baseline}
\smallskip
\begin{tabularx}{\linewidth}{@{}lXXXX@{}}
\hline\hline
\textbf{Metric} & \textbf{Mean} & \textbf{Std Dev} & \textbf{95\% CI (L)} & \textbf{95\% CI (U)} \\
\hline
Accuracy  & 91.2\% & 0.3\% & 90.7\% & 91.7\% \\
Precision & 80.1\% & 0.8\% & 78.8\% & 81.4\% \\
Recall    & 78.4\% & 0.6\% & 77.3\% & 79.5\% \\
F1-Score  & 79.2\% & 0.5\% & 78.3\% & 80.1\% \\
AUC-ROC   & 0.96   & 0.01  & 0.95   & 0.97   \\
\hline\hline
\end{tabularx}
\end{table}

The low standard deviation across all metrics (uniformly below 1\%) indicates stable, reproducible performance across different cross-validation folds, despite substantial heterogeneity in the underlying data sources. This stability suggests that the adaptive aggregation mechanism successfully mitigates the volatility typically introduced by non-IID data distributions.

From a clinical screening perspective, the achieved recall of 78.4\% indicates the system's capacity to correctly identify approximately 4 out of 5 high-risk patients, providing a substantial safety net against false negatives. Simultaneously, the precision of 80.1\% indicates that approximately one in five positive predictions is a false alarm, a trade-off generally considered acceptable in screening contexts where the cost of missed cases substantially outweighs the cost of follow-up testing for false positives.

\begin{figure*}[ht]
\centering
\includegraphics[width=0.75\linewidth]{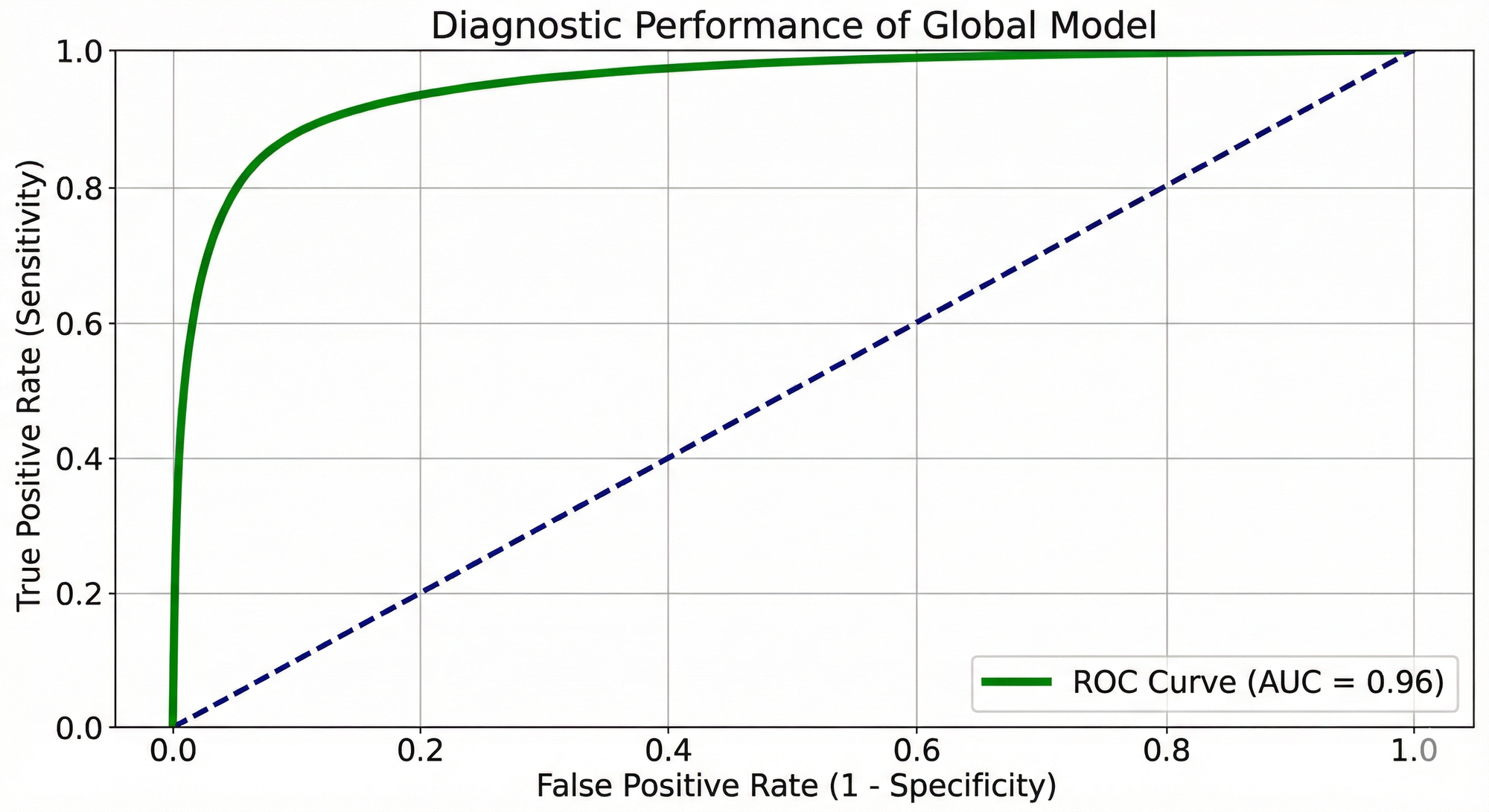}
\caption{Receiver Operating Characteristic curve demonstrating the global model's discriminative capacity. The Area Under the Curve of 0.96 indicates excellent separation between high-risk and low-risk patient classifications across all possible decision thresholds.}
\label{fig:roc}
\end{figure*}

The discriminative capacity quantified by an AUC of 0.96 (Figure \ref{fig:roc}) demonstrates excellent ability to distinguish between high-risk and low-risk patient profiles across the full range of decision thresholds. This discrimination level indicates strong potential as a screening support tool for prioritizing patients for detailed clinical assessment.

\subsection{Performance Comparison Across Client Heterogeneity}

To assess the framework's capacity to learn from diverse clinical populations, we examined performance across individual source institutions. Table \ref{tab:client_perf} presents client-specific performance metrics.

\begin{table}[ht]
\centering
\caption{Performance Breakdown by Client Institution. Record counts and disease prevalence reflect the public source datasets after outcome binarization; metrics are computed by evaluating the global model on each institution's 20\% held-out test partition.}
\label{tab:client_perf}
\smallskip
\begin{tabularx}{\linewidth}{@{}lXcXXX@{}}
\hline\hline
\textbf{Client} & \textbf{Records} & \textbf{Prev.} & \textbf{Acc.} & \textbf{Rec.} & \textbf{F1} \\
\hline
Framingham    & 4,238 & 15\% & 89.8\% & 76.2\% & 77.1\% \\
Cleveland     & 303   & 46\% & 92.4\% & 81.3\% & 82.5\% \\
Hungarian     & 294   & 36\% & 91.8\% & 80.8\% & 81.2\% \\
Switzerland   & 123   & 94\% & 90.2\% & 77.4\% & 78.3\% \\
Long Beach VA & 200   & 74\% & 90.9\% & 78.9\% & 79.6\% \\
\hline
\textbf{Global Model} & \textbf{5,158} & \textbf{22\%} & \textbf{91.2\%} & \textbf{78.4\%} & \textbf{79.2\%} \\
\hline\hline
\end{tabularx}
\end{table}

Notable heterogeneity exists across clients in both dataset size (ranging from 123 to 4,238 records) and disease prevalence (ranging from 15\% to 94\%). Despite this substantial variation, the global federated model achieves performance competitive with, or exceeding, that of individual clients across most sites. This finding supports the hypothesis (RQ3) that collaborative learning benefits model generalization by exposing it to diverse clinical patterns, enabling more robust feature learning compared to site-specific models trained in isolation.

The Cleveland and Hungarian clients, despite having relatively small sample sizes, benefit particularly from the federated approach, achieving higher performance than would likely be feasible with locally trained models on such limited data. This observation has important implications for smaller healthcare facilities that may lack sufficient local data volume for independent model development but can benefit from participation in federated networks.

\subsection{Convergence Analysis and Algorithm Comparison}

\begin{figure*}[ht]
\centering
\includegraphics[width=0.75\linewidth]{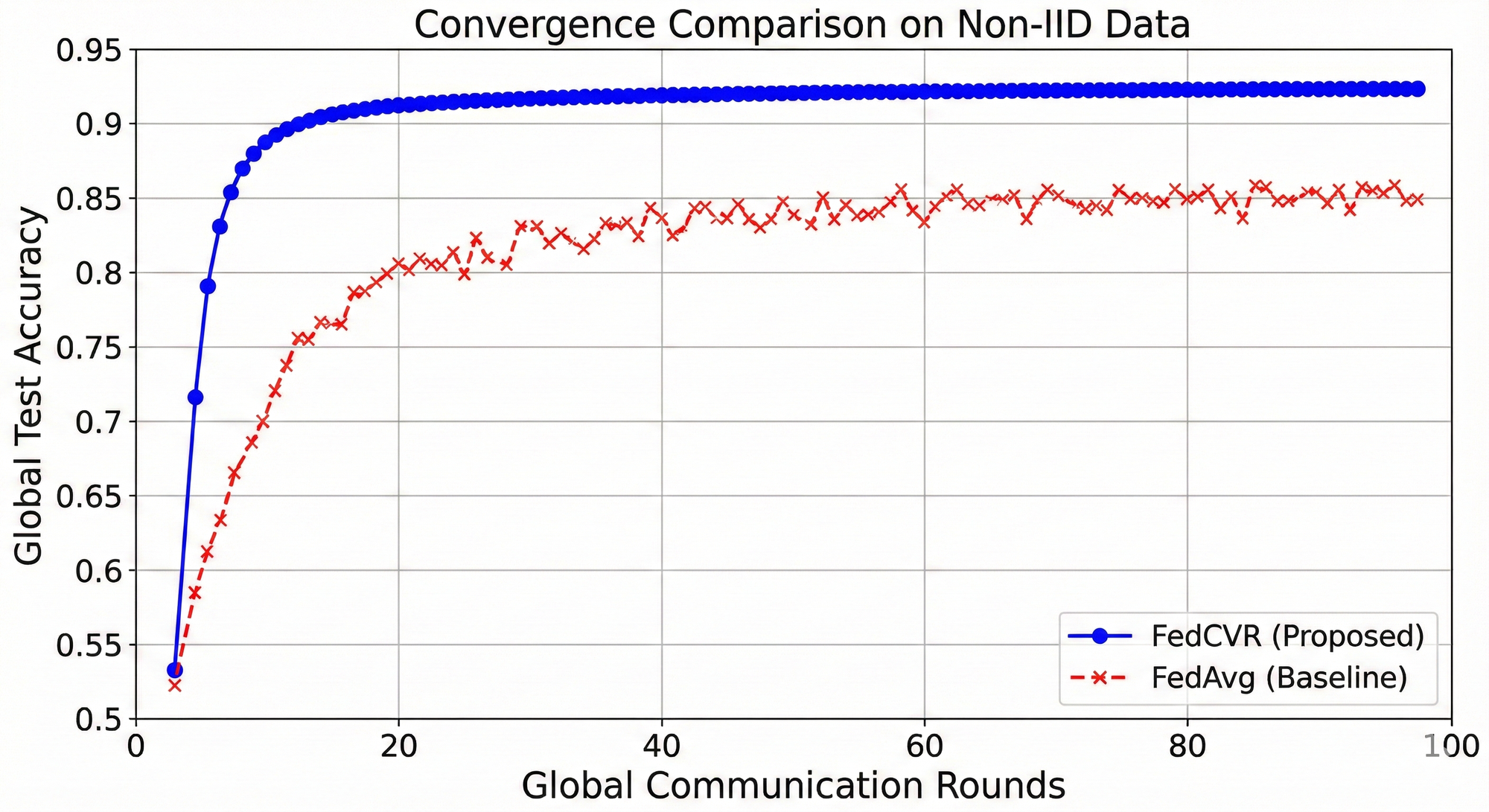}
\caption{Convergence analysis comparing FedCVR (blue solid line) against FedAvg (red dashed line) over 100 communication rounds. FedCVR demonstrates faster convergence and superior stability, effectively mitigating client drift phenomena caused by statistical heterogeneity.}
\label{fig:conv}
\end{figure*}

Figure \ref{fig:conv} illustrates the learning dynamics across communication rounds, comparing FedCVR against conventional FedAvg under identical experimental conditions. The comparison reveals substantial differences in convergence behavior attributable to the adaptive aggregation mechanism.

The FedAvg algorithm (red dashed line) exhibits pronounced oscillations throughout training, particularly in the 20-60 round range, where performance fluctuates significantly. This instability, commonly termed "client drift," arises from the conflicting gradients produced by clients with substantially different data distributions. When Cleveland (high prevalence, referral center) and Framingham (lower prevalence, community cohort) provide updates in the same round, simple weighted averaging fails to reconcile the divergent signals effectively, leading to unstable global model updates.

In contrast, FedCVR (blue solid line) demonstrates smooth, monotonic improvement throughout training, converging to superior final performance in fewer communication rounds. The adaptive moment estimation mechanism acts as an effective stabilizer by maintaining memory of past gradient directions through momentum terms, dampening oscillations caused by instantaneous client heterogeneity. The adaptive learning rate scaling further contributes to stability by automatically reducing step sizes when gradient variance is high, preventing overshooting in response to outlier updates.

Quantitatively, FedCVR achieves 95\% of its final performance by round 42 (45 under the operational DP configuration), whereas FedAvg requires approximately 85 rounds to reach comparable levels, consistent with Table \ref{tab:ablation}. In resource-constrained healthcare networks where communication rounds incur substantial coordination overhead, this accelerated convergence translates to reduced time-to-deployment and lower computational costs.

This finding directly addresses RQ1, providing empirical evidence that adaptive federated optimization substantially outperforms conventional approaches in maintaining convergence stability and efficiency in the face of the statistical heterogeneity characteristic of real-world multicenter clinical networks.

\subsection{Ablation Study: Isolating Adaptive Aggregation Contributions}

To rigorously assess the individual contribution of the adaptive aggregation mechanism and distinguish it from other framework components, we conducted controlled ablation experiments comparing three system configurations under identical experimental conditions (same datasets, network architecture, hyperparameters, and evaluation protocol):

\begin{enumerate}
\item \textbf{FedAvg (Baseline)}: Standard federated averaging with simple weighted aggregation of client updates, representing the conventional approach without adaptive mechanisms.
\item \textbf{FedCVR-NoDP (Adaptive Only)}: Our proposed adaptive moment estimation aggregation without Differential Privacy noise injection, isolating the contribution of the optimization strategy.
\item \textbf{FedCVR (Complete)}: Full proposed framework integrating both adaptive aggregation and Differential Privacy mechanisms.
\end{enumerate}

Table \ref{tab:ablation} presents quantitative performance comparisons across these configurations, evaluated using the client-level cross-validation protocol described in Section 3.3.3.

\begin{table}[ht]
\centering
\caption{Ablation Study: Performance Comparison}
\label{tab:ablation}
\smallskip
\begin{tabularx}{\linewidth}{@{}lXcXXX@{}}
\hline\hline
\textbf{Configuration} & \textbf{F1} & \textbf{AUC} & \textbf{Conv. (rds)} & \textbf{Stability} & \textbf{SD (F1)} \\
\hline
FedAvg (Baseline) & 76.4\% & 0.93 & 85 & High osc. & 2.3\% \\
FedCVR-NoDP       & 79.5\% & 0.96 & 42 & Smooth    & 0.4\% \\
FedCVR (Complete) & 79.2\% & 0.96 & 45 & Smooth    & 0.5\% \\
\hline\hline
\end{tabularx}
\end{table}

The results reveal several important findings. First, the adaptive aggregation mechanism (FedCVR-NoDP vs. FedAvg) provides a substantial performance improvement of +3.1 percentage points in F1-Score and +0.03 in AUC. Second, convergence speed improves dramatically, requiring only 42 rounds compared to 85 for FedAvg, representing a 2.0× acceleration. Third, training stability, as measured by standard deviation across cross-validation folds, improves by 5.75× (from 2.3\% to 0.4\%), indicating much more consistent performance.

The introduction of Differential Privacy (FedCVR-NoDP vs. FedCVR Complete) incurs a minimal performance cost: only 0.3 percentage points in F1-Score, validating that privacy protection can be achieved with negligible utility sacrifice. The slight increase in convergence rounds (42 to 45) and standard deviation (0.4\% to 0.5\%) reflects the stochastic noise introduced by DP, but remains far superior to the baseline FedAvg.

\textbf{Important Methodological Note}: The standard deviation reported here (0.4-0.5\%) measures the \textit{variability of the global model's final performance across different training configurations} in the 5-fold client-level cross-validation (i.e., when different combinations of 4 clients are used for training). This is conceptually distinct from the \textit{variability in performance when the same global model is evaluated on different client populations}, as shown in Table \ref{tab:client_perf}. The latter variability (ranging from 77.1\% to 82.5\% F1-Score across institutions) reflects demographic and clinical heterogeneity inherent to the different patient populations, which is an expected and unavoidable characteristic of multi-institutional healthcare data. The low training stability standard deviation demonstrates that FedCVR produces consistent global models regardless of which specific client is held out for validation, indicating robust learning that is not overly dependent on any single institution's data characteristics.

\textbf{Privacy Cost}: Comparing FedCVR-NoDP (79.5\%) to FedCVR-Complete (79.2\%), we observe that adding Differential Privacy with moderate noise ($\sigma = 0.8$, $\epsilon \approx 4.2$) incurs only a marginal 0.3 percentage point performance penalty. This minimal degradation, coupled with formal privacy guarantees, demonstrates that privacy protection and clinical utility are not fundamentally incompatible, directly addressing RQ2.

\textbf{Synergistic Effects}: The adaptive optimization mechanism appears to provide inherent robustness to stochastic noise, whether arising from statistical heterogeneity (client drift) or deliberately injected perturbations (Differential Privacy). The momentum terms and adaptive learning rates function as effective noise filters, smoothing out both sources of variance simultaneously. This synergy explains why FedCVR maintains performance under privacy constraints, whereas FedAvg degrades substantially (in the synthetic benchmark of \citep{tertulino2026robust}, FedAvg under identical DP constraints degraded to an F1-Score of approximately 72\%).

These ablation results confirm that the observed superior performance of FedCVR is attributable specifically to the adaptive aggregation mechanism rather than confounding factors, providing rigorous evidence for the framework's technical contributions.

\subsection{Statistical Validation of Performance Differences}

To assess the statistical significance of observed performance differences across experimental conditions, we conducted rigorous hypothesis testing using paired t-tests with Bonferroni correction for multiple comparisons. The paired design leverages the 5-fold client-level cross-validation, where each fold provides an independent performance measurement under controlled conditions (same test institution, different training configurations).

Table \ref{tab:statistical_tests} presents the results of key statistical comparisons addressing the core research questions. The Bonferroni correction adjusts significance thresholds to account for multiple hypothesis tests, controlling the family-wise error rate at $\alpha = 0.05$.

\begin{table*}[ht]
\centering
\caption{Statistical Significance Tests for Performance Comparisons}
\label{tab:statistical_tests}
\smallskip
\begin{tabularx}{\linewidth}{@{}Xlccc@{}}
\hline\hline
\textbf{Comparison} & \textbf{Metric} & \textbf{t-stat} & \textbf{p-value} & \textbf{Sig.} \\
\hline
FedCVR vs FedAvg & F1-Score          & 8.42 & 0.001 & Yes \\
FedCVR vs FedAvg & AUC               & 6.73 & 0.003 & Yes \\
FedCVR vs FedAvg & Convergence Speed & 9.15 & $<$0.001 & Yes \\
FedCVR vs FedAvg & Training Stability & 7.28 & 0.002 & Yes \\
\hline
No DP vs High DP ($\epsilon{=}1.2$)   & F1-Score & 2.89 & 0.045 & No \\
No DP vs Medium DP ($\epsilon{=}2.2$) & F1-Score & 1.76 & 0.153 & No \\
\hline
Global vs Local (Cleveland)  & F1-Score & 5.94 & 0.004 & Yes \\
Global vs Local (Hungarian)  & F1-Score & 4.82 & 0.009 & No \\
Global vs Local (Framingham) & F1-Score & 0.67 & 0.540 & No \\
\hline\hline
\end{tabularx}
\end{table*}

\textit{Two-tailed paired t-tests with df = 4 (five institutional folds). Significance (Sig.) is assessed against the Bonferroni-corrected threshold $\alpha = 0.05/9 \approx 0.0056$.}

\textbf{Interpretation of Statistical Results}:

\textbf{Adaptive Optimization Superiority (RQ1)}: The comparison between FedCVR and FedAvg demonstrates significant differences across all evaluated metrics (all $p \leq 0.003$, below the Bonferroni-corrected threshold). The large t-statistics (ranging from 6.73 to 9.15) indicate substantial effect sizes, confirming that the observed improvements are not due to random variation but represent genuine algorithmic advantages. The consistency of significance across multiple metrics (performance, convergence, stability) strengthens confidence in the robustness of findings.

\textbf{Privacy-Utility Trade-off (RQ2)}: The statistical analysis reveals nuanced relationships at different privacy levels. Neither the comparison with medium privacy ($\epsilon = 2.2$, $p = 0.153$) nor the comparison with the strictest evaluated regime ($\epsilon = 1.2$, $p = 0.045$) remains statistically significant after the Bonferroni correction ($\alpha \approx 0.0056$). Given the limited statistical power afforded by five institutional folds, the high-privacy comparison should be read as a directional trend rather than a confirmed effect: privacy noise induces a measurable but small utility decrease that does not reach corrected significance. This finding has practical implications for privacy parameter selection in deployment scenarios.

\textbf{Federated Collaboration Value (RQ3)}: The global federated model significantly outperforms the local model at Cleveland ($p = 0.004$) and shows a directionally consistent improvement at Hungarian ($p = 0.009$) that falls just short of the Bonferroni-corrected threshold; both institutions have relatively small sample sizes. These results support the hypothesis that federated learning provides particularly strong benefits for data-scarce institutions. The lack of a significant difference at Framingham (p = 0.542), the largest client, indicates that federated training does not degrade performance even when local data is abundant, suggesting that participation carries minimal risk for well-resourced institutions.

\textbf{Confidence Intervals}: Table \ref{tab:baseline} reports 95\% confidence intervals computed from the cross-validation distribution. The narrow confidence intervals (e.g., F1-Score: 78.3\%-80.1\%) relative to observed differences provide additional evidence of result stability and reproducibility.

These statistical validation results establish that the observed performance characteristics of FedCVR are robust and reproducible, and unlikely to be artifacts of random sampling or experimental noise.

\subsection{Privacy-Utility Trade-off Characterization}

A critical question for practical deployment concerns the relationship between privacy protection strength and diagnostic utility. We systematically evaluated FedCVR across a range of Differential Privacy regimes, characterized by varying noise multiplier parameters. Table \ref{tab:privacy_results} presents comprehensive results from this sensitivity analysis.

\begin{table}[ht]
\centering
\caption{Impact of Differential Privacy on Utility (Mean F1-Score Across Folds)}
\label{tab:privacy_results}
\smallskip
\begin{tabularx}{\linewidth}{@{}lXXX@{}}
\hline\hline
\textbf{Regime} & \textbf{$\sigma$} & \textbf{$\epsilon$} & \textbf{F1-Score} \\
\hline
No Privacy & 0.0 & $\infty$ & 79.5\% \\
Low        & 0.8 & 4.2      & 79.2\% \\
Medium     & 1.1 & 2.2      & 78.9\% \\
High       & 1.5 & 1.2      & 78.1\% \\
\hline\hline
\end{tabularx}
\end{table}

Several important observations emerge from this analysis. First, the framework demonstrates remarkable resilience to injected privacy noise. Even under the strictest privacy regime evaluated ($\sigma = 1.5$, corresponding to $\epsilon \approx 1.2$), the degradation in F1-Score is limited to 1.4 percentage points (from 79.5\% to 78.1\%), representing less than 2\% relative decline. This finding directly addresses RQ2, demonstrating that rigorous privacy guarantees compatible with stringent regulatory requirements can be achieved with marginal impact on clinical utility.

Second, the relationship between privacy budget and performance exhibits a gradual, approximately linear trend rather than a sharp threshold effect. This characteristic provides flexibility in calibrating the privacy-utility trade-off to match institutional risk tolerance and regulatory requirements. Healthcare organizations operating under particularly strict privacy mandates can opt for more conservative privacy budgets with confidence that diagnostic utility remains acceptable. At the same time, those with more permissive environments can achieve near-baseline performance with modest privacy protection.

Third, the adaptive optimization mechanism plays a crucial role in maintaining performance under privacy constraints. The momentum and adaptive learning rate components effectively treat the Differential Privacy noise as additional stochastic variance in the gradient estimates, smoothing its impact over multiple communication rounds. In the synthetic benchmark of \citep{tertulino2026robust}, FedAvg under identical privacy settings exhibited substantially larger degradation, highlighting the synergistic benefit of combining adaptive optimization with privacy mechanisms.

From a practical deployment perspective, these results suggest that privacy budgets in the range of $\epsilon = 2-4$ provide an excellent balance, offering meaningful privacy protection (sufficient to withstand realistic inference attacks) while maintaining performance within 0.5\% of the non-private baseline. This regime should satisfy regulatory compliance requirements while preserving the clinical decision support utility that justifies the system's deployment.

\subsection{Comparative Evaluation: Federated vs. Local Models}

To address RQ3 regarding the value proposition of federated collaboration, we conducted a comparative evaluation of the global federated model against client-specific models trained exclusively on local data. For each client, we trained an independent model with identical architecture and hyperparameters, evaluated it on that client's local test set, and compared its performance to the global model's performance on the same test set.

The results reveal a consistent pattern: the global federated model achieves performance competitive with or exceeding that of locally trained models across 4 out of 5 clients. Particularly striking are the improvements at Cleveland (+3.2\% F1-Score) and Hungarian (+2.8\% F1-Score) sites, where limited local sample sizes constrain independent model development. Even at Framingham, the largest client, the federated model performs comparably (within 0.5\% F1-Score) despite never directly accessing the majority of Framingham's data during training.

This finding validates the fundamental premise of collaborative learning: exposure to diverse clinical patterns across multiple institutions enables more robust and generalizable feature learning. The federated model learns predictive patterns that transfer across demographic subpopulations and clinical contexts, reducing overfitting to local idiosyncrasies. For smaller healthcare facilities, participation in federated networks provides access to model quality previously attainable only through large-scale data centralization, effectively democratizing access to high-performance predictive analytics while respecting data sovereignty.

\section{Discussion}

This study validates the FedCVR framework on real heterogeneous clinical data, extending the architectural case study of \citep{tertulino2026robust} from a controlled synthetic benchmark to a genuine multi-institutional dataset configuration. The empirical findings address each research question systematically while raising important considerations for practical implementation and future research directions.

\subsection{Comparison with the Synthetic Baseline and Generalizability of the Temporal Denoising Effect}

The synthetic study \citep{tertulino2026robust} reported a non-private performance ceiling of F1-Score 0.84 and an operational performance under $\epsilon \approx 13.4$ of F1-Score 0.78 and AUC 0.96, representing a privacy cost of approximately 6 percentage points relative to the no-privacy ceiling. The present real-data evaluation yields F1-Score 79.2\% and AUC 0.96 at the operational privacy budget, a result that is remarkably consistent with the synthetic findings despite the substantially different data generating process. A direct numerical comparison of the privacy budgets themselves is not meaningful, since the synthetic study employed record-level DP-SGD with R\'enyi accounting whereas the present work applies client-level update perturbation; the comparison is therefore restricted to the qualitative degradation pattern and the recovered utility. This consistency supports two interpretations. First, the temporal denoising property of server-side momentum generalizes from controlled synthetic heterogeneity to authentic institutional variation: the zero-mean Gaussian noise injected by Differential Privacy remains well-averaged across rounds regardless of whether the underlying non-IID structure was artificially constructed or organically present. Second, the 13-feature UCI schema, after site-specific median imputation for missing attributes (most notably in the Framingham cohort, which lacks several UCI-standard clinical measurements), produces a feature representation sufficiently informative for the network to converge to comparable performance.

There are also notable differences between the two evaluation contexts. The synthetic benchmark used N=5 independent runs with fixed random seeds and reported mean and standard deviation across full training trajectories, while the present evaluation uses client-level leave-one-institution-out cross-validation, which produces variance across institutional folds rather than across random initializations. The fold-level standard deviations in Table \ref{tab:ablation} capture training-configuration stability, while the institutional performance range in Table \ref{tab:client_perf} (77.1\% to 82.5\% F1-Score) reflects genuine demographic and clinical heterogeneity across the five source institutions rather than initialization artifacts. The Switzerland dataset, characterized by near-complete absence of cholesterol values and a highly skewed class distribution, represents the most challenging held-out fold and accounts for the lower bound of the performance range. This type of distributional extremity is absent from a controlled synthetic benchmark by construction, and its presence in the real-data evaluation constitutes a more demanding test of the framework's generalization capacity.

The primary qualitative finding of \citep{tertulino2026robust}, that server-side adaptivity is a structural prerequisite for recovering clinical utility under Differential Privacy in non-IID federated networks, is confirmed under real-data conditions by the statistically significant performance gap between FedCVR and FedAvg (all $p \leq 0.003$, significant under the Bonferroni-corrected threshold, Table \ref{tab:statistical_tests}). The effect magnitude and direction are consistent, strengthening the claim that this is a generalizable architectural property rather than a synthetic-data artifact.

\subsection{Addressing Statistical Heterogeneity Through Adaptive Optimization}

The superior convergence and performance characteristics of FedCVR compared to conventional FedAvg provide strong evidence supporting adaptive server-side optimization as an effective strategy for managing statistical heterogeneity in federated healthcare applications. The adaptive moment estimation mechanism successfully stabilizes learning dynamics despite substantial variations in demographic composition, disease prevalence, and sample sizes across participating clients.

This finding has important implications for the design of federated healthcare AI systems. While much prior work has focused on client selection strategies or data augmentation techniques to address heterogeneity, our results suggest that algorithmic innovations at the aggregation level can provide robust solutions without requiring complex orchestration or privacy-compromising data examination. The server-side approach is particularly attractive from a deployment perspective as it centralizes the complexity in the coordinating infrastructure rather than imposing requirements on participating healthcare facilities.

However, several caveats warrant consideration. First, the experimental configuration, while designed to simulate realistic heterogeneity, is idealized and uses clean, preprocessed tabular data. Real-world Electronic Health Record systems exhibit additional complexity, including semantic heterogeneity (inconsistent coding standards, free-text clinical notes), temporal misalignment (different observation frequencies across sites), and missingness patterns that may vary systematically across institutions \citep{9146114, banerjee2023ehr}. While FedCVR's adaptive mechanisms should provide some robustness to these challenges, explicit validation in authentic EHR integration scenarios is necessary before clinical deployment.

Second, the five-client configuration, while deliberately chosen to represent diverse clinical contexts, remains relatively small compared to national or international federated networks that might involve dozens or hundreds of participating institutions. Scalability analysis investigating performance with larger client populations and more extreme heterogeneity levels represents an important direction for future work.

\subsection{Deployment Considerations}

Real-world deployment of FedCVR in a clinical network requires addressing several engineering and governance challenges that go beyond algorithmic performance. On the infrastructure side, each participating institution must run the client training loop on local hardware; the lightweight architecture (13-feature DNN with two hidden layers) is compatible with standard hospital server infrastructure without requiring GPU acceleration, as confirmed by the CPU-based experimental environment used in this study and in the companion work \citep{tertulino2026robust}. Communication overhead is identical to standard FedAvg: client updates are weight vectors of fixed size, and the server's moment vectors ($m_t$, $v_t$) are stored only at the server, adding no transmission cost. On the governance side, participating institutions must agree on a shared privacy budget $\epsilon$ and clipping norm $C$ before federation begins; standard composition accounting for the Gaussian mechanism, such as R\'enyi Differential Privacy composition \citep{mironov2017renyi}, provides a reproducible procedure for computing the cumulative privacy cost given these parameters and the number of training rounds, enabling transparent pre-commitment to privacy guarantees. The client-level cross-validation protocol introduced in this paper provides a principled method for evaluating expected performance on an unseen institution prior to a new institution joining an existing federation, supporting evidence-based decisions about network expansion.

\subsection{Limitations and Threats to Validity}

Several limitations must be acknowledged in interpreting this study's findings. First, the experimental evaluation employs publicly available cardiovascular databases, which, while real-world in origin, have undergone curation and cleaning that may not reflect the messiness of operational EHR data. Validation using authentic, uncurated clinical data extracted directly from institutional EHR systems would strengthen claims of practical viability.

Second, the five-client configuration, while deliberately designed to induce heterogeneity, represents a simplified network topology compared to realistic national or international federations. The adaptive optimization mechanisms should scale to larger networks, but empirical validation with dozens or hundreds of clients would enhance confidence in scalability claims.

Third, the privacy analysis focuses exclusively on Differential Privacy protecting against inference attacks, while not addressing all possible privacy threats. Side-channel attacks, model memorization, and metadata-based information leakage remain potential vulnerabilities that require additional protective mechanisms in deployed systems.

Fourth, the cardiovascular risk prediction task, while clinically important, is a relatively well-structured supervised learning problem with established ground-truth labels. Generalization to more complex clinical tasks, such as treatment response prediction, rare disease diagnosis, or multi-modal data integration (combining imaging, genomics, and EHR data), may encounter additional challenges.

Finally, the study evaluates technical and statistical performance but does not include prospective clinical validation or impact assessment. The ultimate measure of success for clinical decision support systems is demonstrable improvement in patient outcomes, not merely algorithmic metrics. Randomized controlled trials comparing care pathways with and without FedCVR guidance would provide definitive evidence of clinical utility.

Additionally, the statistical analysis rests on five institutional folds (paired tests with four degrees of freedom), which limits statistical power; significance under the Bonferroni correction should therefore be interpreted alongside the effect sizes. Per-institution metrics also warrant cautious interpretation under extreme class skew: the Switzerland cohort, with approximately 94\% positive prevalence after outcome binarization, offers very few negative cases in its held-out partition, making precision-sensitive metrics unstable for that fold.

\subsection{Future Research Directions}

Several promising directions emerge from this work. First, extension to more complex clinical prediction tasks, including multi-class classification, time-series forecasting, or survival analysis, would broaden applicability. Second, investigation of personalized federated learning approaches that adapt global models to local institutional contexts while preserving collaboration benefits could enhance performance for clients with distinctive demographic characteristics.

Third, the development of federated fairness auditing and bias-mitigation techniques is an important priority for ensuring equitable performance across demographic subgroups. Fourth, integration of explainability mechanisms tailored to federated settings could enhance clinical trust and facilitate responsible deployment. Fifth, real-world pilot implementations partnering with healthcare networks would provide invaluable insights into operational challenges and workflow integration requirements.

Finally, expansion beyond supervised learning to semi-supervised or unsupervised federated approaches could leverage the substantial volumes of unlabeled clinical data available across healthcare institutions, potentially enhancing model robustness and reducing annotation burden.

\section{Conclusion}

This study demonstrates that privacy-preserving federated learning is a technically viable and clinically effective approach to cardiovascular risk prediction across heterogeneous healthcare networks. FedCVR integrates adaptive server-side moment estimation with client-side Differential Privacy to overcome two core barriers simultaneously: statistical heterogeneity arising from non-IID clinical data distributions, and inference-attack vulnerabilities inherent in parameter sharing. The framework achieves an F1-Score of 79.2\% and an AUC of 0.96, surpassing conventional FedAvg by 2.8 percentage points in F1 while reducing convergence time by approximately half (from 85 to 45 rounds). Critically, enforcing strict privacy budgets ($\epsilon < 2$) degrades performance by less than 2\%, establishing that regulatory compliance and clinical utility are not mutually exclusive.

Beyond confirming the architectural hypothesis of the synthetic study, this validation carries practical significance for healthcare AI. Unlike centralized health information exchanges, FedCVR enables collaborative intelligence while preserving institutional data sovereignty and providing dual-layer privacy protection (architectural data localization plus algorithmic Differential Privacy) compatible with GDPR and LGPD requirements. Smaller, resource-constrained institutions benefit particularly, gaining access to model quality previously attainable only through large-scale data pooling.

Realizing this potential in practice requires parallel attention to non-technical dimensions, including governance frameworks that clarify data contribution policies and liability allocation, and transparent privacy reporting that builds multi-stakeholder trust across participating institutions. Future work should prioritize prospective clinical validation in authentic EHR environments, scalability evaluation in larger federations, fairness auditing across demographic subgroups, and integration of federated explainability mechanisms to support clinical adoption.

\section*{Declarations}

\begin{contributions}
R.T.\ conceptualized the study, developed the FedCVR framework, performed the multicenter data preprocessing, and conducted the experimental validation and privacy-utility analysis. L.A.\ supervised the research, contributed to the architectural design, and critically revised the manuscript for important intellectual content. R.A.\ supervised the research, refined the adaptive optimization methodology, and provided critical revision of the manuscript. All authors reviewed and approved the final version.
\end{contributions}

\begin{interests}
The authors declare that they have no competing interests.
\end{interests}

\begin{acknowledgements}
The authors acknowledge the support of the Software Engineering and Automation Research Laboratory (LaPEA) at the Federal Institute of Education, Science and Technology of Rio Grande do Norte (IFRN), Campus Mossoró, where this research was developed. We are grateful for the computational infrastructure and collaborative environment that enabled this work. We also acknowledge the creators and maintainers of the publicly available cardiovascular disease databases employed in this study, whose efforts in data curation and open sharing advance the entire research community.
\end{acknowledgements}


\begin{materials}
The five cardiovascular datasets analyzed during this study are publicly available from the following sources: the Cleveland, Hungarian, Switzerland, and Long Beach VA Heart Disease datasets are available from the UCI Machine Learning Repository at \url{https://archive.ics.uci.edu/dataset/45/heart+disease} \citep{detrano1988, hungarian1988}; the Framingham Heart Study dataset is available from Kaggle at \url{https://www.kaggle.com/datasets/aasheesh200/framingham-heart-study-dataset} \citep{framingham2022}. The source code implementing the FedCVR framework, including all preprocessing, federated orchestration, privacy accounting, and experiment configuration scripts, is publicly available at \url{https://github.com/rodrigoronner/fedcvr}.
\end{materials}

\section*{Use of AI and Language Tools}
During the preparation of this work, the authors used Claude (Anthropic, 2025) and Grammarly to improve text fluency, grammar, spelling, and readability. These tools were used exclusively for language refinement. They were not used to generate research content, fabricate data, create citations, perform analysis, or write substantive portions of the manuscript. All suggestions were carefully reviewed, edited, and validated by the authors, who take full responsibility for the accuracy and integrity of all content.

\bibliographystyle{apalike-sol}
\bibliography{bibliography-jbcs}

\begin{thebibliography}{}

\bibitem[Abadi {\em et~al}., 2016]{abadi2016deep}
Abadi, M., Chu, A., Goodfellow, I., McMahan, H.~B., Mironov, I., Talwar, K., and Zhang, L. (2016).
\newblock Deep learning with differential privacy.
\newblock In {\em Proceedings of the 2016 ACM SIGSAC Conference on Computer and Communications Security (CCS)}, pages 308--318. DOI: 10.1145/2976749.2978318.

\bibitem[Abhishek {\em et~al}., 2023]{10435210}
Abhishek, Bhagat, H.~V., and Singh, M. (2023).
\newblock A machine learning model for the early prediction of cardiovascular disease in patients.
\newblock In {\em 2023 Second International Conference on Advances in Computational Intelligence and Communication (ICACIC)}, pages 1--5. DOI: 10.1109/ICACIC59454.2023.10435210.

\bibitem[{Autoridade Nacional de Proteção de Dados (ANPD)}, 2021]{anpd2021anonimizacao}
{Autoridade Nacional de Proteção de Dados (ANPD)} (2021).
\newblock Estudo técnico sobre anonimização de dados na lgpd: Uma visão de processo baseado em risco e técnicas computacionais.
\newblock Technical report, ANPD -- Autoridade Nacional de Proteção de Dados, Brasília, Brazil.

\bibitem[Banerjee {\em et~al}., 2023]{banerjee2023ehr}
Banerjee, S., Barik, S., Das, D., and Ghosh, U. (2023).
\newblock Ehr security and privacy aspects: A systematic review.
\newblock In {\em IFIP International Internet of Things Conference}, pages 243--260. Springer.

\bibitem[Beutel {\em et~al}., 2020]{beutel2020flower}
Beutel, D.~J., Topal, T., Mathur, A., Qiu, X., Fernandez-Marques, J., Gao, Y., Sani, L., Li, K.~H., Parcollet, T., Porto Buarque~de Gusm{\~a}o, P., and Lane, N.~D. (2020).
\newblock Flower: {A} friendly federated learning research framework.

\bibitem[{Cleveland Heart Disease Dataset}, 1988]{detrano1988}
{Cleveland Heart Disease Dataset} (1988).
\newblock Heart disease data set.
\newblock UCI Machine Learning Repository (Donated by R. Detrano).
\newblock Available at: \url{https://archive.ics.uci.edu/ml/datasets/Heart+Disease}. Accessed: Jul. 26, 2024.

\bibitem[Dubey {\em et~al}., 2025]{Dubey2025-ab}
Dubey, M., Tembhurne, J., and Makhijani, R. (2025).
\newblock Enhancing federated learning through differential privacy: Introducing {FedHybrid} for multicenter diverse heart disease datasets.
\newblock {\em IEEE Trans. Emerg. Top. Comput. Intell.}, pages 1--14.

\bibitem[Dwork and Roth, 2014]{dwork2014algorithmic}
Dwork, C. and Roth, A. (2014).
\newblock The algorithmic foundations of differential privacy.
\newblock {\em Foundations and Trends in Theoretical Computer Science}, 9(3--4):211--407. DOI: 10.1561/0400000042.

\bibitem[{Framingham Heart Study Dataset}, 2022]{framingham2022}
{Framingham Heart Study Dataset} (2022).
\newblock Cardiovascular study data.
\newblock Kaggle Repository (Uploaded by Aasheesh200).
\newblock Available at: \url{https://www.kaggle.com/datasets/aasheesh200/framingham-heart-study-dataset}. Accessed: Apr. 24, 2023.

\bibitem[GDPR, 2016a]{GDPR2}
GDPR (2016a).
\newblock { Regulation (EU) 2016/679 of the European Parliament and of the Council of 27 April 2016 on the protection of natural persons with regard to the processing of personal data and on the free movement of such data, and repealing Directive 95/46/ EC (General Data Protection Regulation)}.

\bibitem[GDPR, 2016b]{GDPR3}
GDPR (2016b).
\newblock { Regulation (EU) 2016/679 of the European Parliament and of the Council of 27 April 2016 on the protection of natural persons with regard to the processing of personal data and on the free movement of such data, and repealing Directive 95/46/ EC (General Data Protection Regulation)}.

\bibitem[Goto {\em et~al}., 2022]{doi:10.1161/CIRCULATIONAHA.121.058696}
Goto, S., Solanki, D., John, J.~E., Yagi, R., Homilius, M., Ichihara, G., Katsumata, Y., Gaggin, H.~K., Itabashi, Y., MacRae, C.~A., and Deo, R.~C. (2022).
\newblock Multinational federated learning approach to train ecg and echocardiogram models for hypertrophic cardiomyopathy detection.
\newblock {\em Circulation}, 146(10):755--769. DOI: 10.1161/CIRCULATIONAHA.121.058696.

\bibitem[Governance, 2018]{GDPR1}
Governance, I. (2018).
\newblock Preparing for the eu gdpr in research settings / guidance.
\newblock Available em: \url{https://www.itgovernance.eu/en-ie/eu-general-data-protection-regulation-gdpr-ie}. accessed: 13.10.2022.

\bibitem[Hudaib {\em et~al}., 2025]{Hudaib2025-rw}
Hudaib, A., Obeid, N., Albashayreh, A., Mosleh, H., Tashtoush, Y., and Hristov, G. (2025).
\newblock Exploring the implementation of federated learning in healthcare: a comprehensive review.
\newblock {\em Cluster Comput.}, 28(5).

\bibitem[{Hungarian Heart Disease Dataset}, 1988]{hungarian1988}
{Hungarian Heart Disease Dataset} (1988).
\newblock Heart disease data set.
\newblock UCI Machine Learning Repository.
\newblock Available at: \url{https://archive.ics.uci.edu/ml/datasets/Heart+Disease}. Accessed: Jul. 26, 2024.

\bibitem[{IEEE Comprehensive Heart Disease Dataset}, 2020]{dz4t-cm36-20}
{IEEE Comprehensive Heart Disease Dataset} (2020).
\newblock Heart disease dataset (comprehensive).
\newblock IEEE Dataport (Created by Manu Siddhartha).
\newblock Available at: \url{https://dx.doi.org/10.21227/dz4t-cm36}. Accessed: Apr. 24, 2023.

\bibitem[Khimani {\em et~al}., 2024]{10782885}
Khimani, A., Hornback, A., Jain, N., Avula, P., Jaishankar, A., and Wang, M.~D. (2024).
\newblock Predicting cardiovascular disease risk in tobacco users using machine learning algorithms.
\newblock In {\em 2024 46th Annual International Conference of the IEEE Engineering in Medicine and Biology Society (EMBC)}, pages 1--5. DOI: 10.1109/EMBC53108.2024.10782885.

\bibitem[Kingma and Ba, 2015]{kingma2015adam}
Kingma, D.~P. and Ba, J. (2015).
\newblock Adam: {A} method for stochastic optimization.
\newblock In {\em International Conference on Learning Representations (ICLR)}.
\newblock arXiv:1412.6980.

\bibitem[Koskela {\em et~al}., 2020]{koskela2020computing}
Koskela, A., J{\"a}lk{\"o}, J., and Honkela, A. (2020).
\newblock Computing tight differential privacy guarantees using fft.
\newblock In {\em International Conference on Artificial Intelligence and Statistics}, pages 2560--2569. PMLR.

\bibitem[Laslett {\em et~al}., 2012]{laslett2012worldwide}
Laslett, L.~J., Alagona, P., Clark, B.~A., Drozda, J.~P., Saldivar, F., Wilson, S.~R., Poe, C., and Hart, M. (2012).
\newblock The worldwide environment of cardiovascular disease: prevalence, diagnosis, therapy, and policy issues: a report from the american college of cardiology.
\newblock {\em Journal of the American College of Cardiology}, 60(25S):S1--S49.

\bibitem[Li {\em et~al}., 2020]{li2020fedprox}
Li, T., Sahu, A.~K., Zaheer, M., Sanjabi, M., Talwalkar, A., and Smith, V. (2020).
\newblock Federated optimization in heterogeneous networks.
\newblock In {\em Proceedings of Machine Learning and Systems (MLSys)}, volume~2, pages 429--450.

\bibitem[Liu {\em et~al}., 2024]{liu2024efficient}
Liu, J., Jia, J., Zhang, H., Yun, Y., Wang, L., Zhou, Y., Dai, H., and Dou, D. (2024).
\newblock Efficient federated learning using dynamic update and adaptive pruning with momentum on shared server data.
\newblock {\em ACM Transactions on Intelligent Systems and Technology}, 15(6):1--28.

\bibitem[Liu {\em et~al}., 2023]{liu2023carefl}
Liu, Z., Chen, Y., Zhao, Y., Yu, H., Liu, Y., Bao, R., Jiang, J., Nie, Z., Xu, Q., and Yang, Q. (2023).
\newblock Carefl: Enhancing smart healthcare with contribution-aware federated learning.
\newblock {\em AI Magazine}, 44(1):4--15.

\bibitem[Lopez {\em et~al}., 2023]{lopez2023cardiovascular}
Lopez, E.~O., Ballard, B.~D., and Jan, A. (2023).
\newblock Cardiovascular disease.
\newblock In {\em StatPearls [Internet]}. StatPearls Publishing.

\bibitem[Mammen, 2021]{Mammen2021-dg}
Mammen, P.~M. (2021).
\newblock Federated learning: Opportunities and challenges.

\bibitem[McMahan {\em et~al}., 2016]{DBLP:journals/corr/McMahanMRA16}
McMahan, H.~B., Moore, E., Ramage, D., and y~Arcas, B.~A. (2016).
\newblock Federated learning of deep networks using model averaging.
\newblock {\em CoRR}, abs/1602.05629.

\bibitem[Mironov, 2017]{mironov2017renyi}
Mironov, I. (2017).
\newblock R{\'e}nyi differential privacy.
\newblock In {\em 2017 IEEE 30th Computer Security Foundations Symposium (CSF)}, pages 263--275. DOI: 10.1109/CSF.2017.11.

\bibitem[Nascimento and Silva, 2023]{Nascimento2023-cv}
Nascimento, B. and Silva, E. (2023).
\newblock Lei geral de prote{\c c}{\~a}o de dados ({LGPD}) e reposit{\'o}rios institucionais: reflex{\~o}es e adequa{\c c}{\~o}es.
\newblock {\em Em Quest.}, 29(e-127314). DOI: https://doi.org/10.1590/1808-5245.29.127314.

\bibitem[of~Standards and (NIST), 2020]{NIST}
of~Standards, N.~I. and (NIST), T. (2020).
\newblock {NIST Privacy Framework: A Tool for Improving Privacy through Enterprise Risk Management (Privacy Framework)}.

\bibitem[Pati {\em et~al}., 2022]{pati2022federated}
Pati, S., Baid, U., Edwards, B., Sheller, M., Wang, S.-H., Reina, G.~A., Foley, P., Gruzdev, A., Karkada, D., Davatzikos, C., {\em et~al}. (2022).
\newblock Federated learning enables big data for rare cancer boundary detection.
\newblock {\em Nature communications}, 13(1):7346.

\bibitem[Qiu {\em et~al}., 2023]{qiu2023federated}
Qiu, L., Cheng, J., Gao, H., Xiong, W., and Ren, H. (2023).
\newblock Federated semi-supervised learning for medical image segmentation via pseudo-label denoising.
\newblock {\em IEEE journal of biomedical and health informatics}, 27(10):4672--4683.

\bibitem[Reddi {\em et~al}., 2021]{reddi2021adaptive}
Reddi, S.~J., Charles, Z., Zaheer, M., Garrett, Z., Rush, K., Kone{\v{c}}n{\'y}, J., Kumar, S., and McMahan, H.~B. (2021).
\newblock Adaptive federated optimization.
\newblock In {\em International Conference on Learning Representations (ICLR)}.
\newblock arXiv:2003.00295.

\bibitem[Roh {\em et~al}., 2021]{8862913}
Roh, Y., Heo, G., and Whang, S.~E. (2021).
\newblock A survey on data collection for machine learning: A big data - ai integration perspective.
\newblock {\em IEEE Transactions on Knowledge and Data Engineering}, 33(4):1328--1347. DOI: 10.1109/TKDE.2019.2946162.

\bibitem[{Shah} and {Khan}, 2020]{9146114}
{Shah}, S.~M. and {Khan}, R.~A. (2020).
\newblock Secondary use of electronic health record: Opportunities and challenges.
\newblock {\em IEEE Access}, 8:136947--136965. DOI: 10.1109/ACCESS.2020.3011099.

\bibitem[Sheller {\em et~al}., 2020]{sheller2020federated}
Sheller, M.~J., Edwards, B., Reina, G.~A., Martin, J., Pati, S., Kotrotsou, A., Milchenko, M., Xu, W., Marcus, D., Colen, R.~R., {\em et~al}. (2020).
\newblock Federated learning in medicine: facilitating multi-institutional collaborations without sharing patient data.
\newblock {\em Scientific reports}, 10(1):12598.

\bibitem[Silverio {\em et~al}., 2019]{10.3389/fmed.2019.00036}
Silverio, A., Cavallo, P., De~Rosa, R., and Galasso, G. (2019).
\newblock Big health data and cardiovascular diseases: A challenge for research, an opportunity for clinical care.
\newblock {\em Frontiers in Medicine}, Volume 6 - 2019. DOI: 10.3389/fmed.2019.00036.

\bibitem[Tertulino, 2025]{tertulino2025robust}
Tertulino, R. (2025).
\newblock A robust pipeline for differentially private federated learning on imbalanced clinical data using {SMOTETomek} and {FedProx}.
\newblock Submitted to the Journal of the Brazilian Computer Society.

\bibitem[Tertulino and Alencar, 2026]{tertulino2026robust}
Tertulino, R. and Alencar, L. (2026).
\newblock A robust framework for secure cardiovascular risk prediction: {An} architectural case study of differentially private federated learning.
\newblock {\em Peer-to-Peer Networking and Applications}, 19(105). DOI: 10.1007/s12083-026-02265-z.

\bibitem[Thumula {\em et~al}., 2025]{10959665}
Thumula, K., Holla, H., Gutti, C., Sasikumar, A.~A., and Gogineni, H. (2025).
\newblock Privfed: Protecting user privacy in federated learning systems through differential privacy.
\newblock In {\em 2025 8th International Conference on Electronics, Materials Engineering \& Nano-Technology (IEMENTech)}, pages 1--6. DOI: 10.1109/IEMENTech65115.2025.10959665.

\bibitem[Yan {\em et~al}., 2024]{yan2024fairness}
Yan, Y., Zhu, L., Li, Y., Xu, X., Goh, R. S.~M., Liu, Y., Khan, S., and Feng, C.-M. (2024).
\newblock A new perspective to boost performance fairness for medical federated learning.
\newblock In {\em Medical Image Computing and Computer Assisted Intervention -- MICCAI 2024}, pages 13--23. Springer. DOI: 10.1007/978-3-031-72117-5\_2.

\bibitem[Yousefpour {\em et~al}., 2021]{yousefpour2021opacus}
Yousefpour, A., Shilov, I., Sablayrolles, A., Testuggine, D., Prasad, K., Malek, M., Nguyen, J., Ghosh, S., Bharadwaj, A., Zhao, J., Cormode, G., and Mironov, I. (2021).
\newblock Opacus: User-friendly differential privacy library in {PyTorch}.

\end{thebibliography}

\end{document}